\pdfoutput=1

\documentclass[11pt]{article}

\usepackage[]{ACL2023}

\usepackage{times}
\usepackage{latexsym}

\usepackage[T1]{fontenc}

\usepackage[utf8]{inputenc}

\usepackage{microtype}
\usepackage{inconsolata}
\usepackage{amsfonts,amsmath}
\usepackage{booktabs}
\usepackage{multirow}
\usepackage{graphicx}
\usepackage{bm}
\usepackage{subfigure}
\usepackage{enumerate}
\usepackage{enumitem}
\usepackage{bbding}
\usepackage[nameinlink]{cleveref}

\usepackage{algorithm}
\usepackage{algpseudocode}
\usepackage{mathtools}
\usepackage{arydshln}
\usepackage{colortbl}

\usepackage{caption}

\usepackage[nameinlink]{cleveref}
\crefformat{section}{\S#2#1#3} 
\crefname{algorithm}{Alg.}{Algs.}
\crefformat{subsection}{\S#2#1#3}
\Crefname{equation}{Eq.}{Eqs.}
\Crefname{figure}{Fig.}{Figs.}

\title{KGA: A General Machine Unlearning Framework Based on \\ Knowledge Gap Alignment}

\author{Lingzhi Wang$^{1,2}$, Tong Chen$^3$, Wei Yuan$^{3}$, Xingshan Zeng$^4$, Kam-Fai Wong$^{1,2}$, Hongzhi Yin$^3$\\
  $^1$The Chinese University of Hong Kong, Hong Kong, China\\
  $^2$MoE Key Laboratory of High Confidence Software Technologies, China\\
  $^3$School of Information Technology
and Electrical Engineering, The University of Queensland\\
  \tt $^{1,2}$\{lzwang,kfwong\}@se.cuhk.edu.hk \\
   \tt $^3$\{tong.chen,w.yuan,h.yin1\}uq.edu.au, \tt $^4$zxshamson@gmail.com
}

\begin{document}
\maketitle

\begin{abstract}
Recent legislation of the ``right to be forgotten'' has led to the interest in machine unlearning, where the learned models are endowed with the function to forget information about specific training instances as if they have never existed in the training set. Previous work mainly focuses on computer vision scenarios and largely ignores the essentials of unlearning in NLP field, where text data contains more explicit and sensitive personal information than images. In this paper, we propose a general unlearning framework called KGA to induce forgetfulness. Different from previous work that tries to recover gradients or forces models to perform close to one specific distribution, KGA maintains distribution differences (i.e., knowledge gap). This relaxes the distribution assumption. Furthermore, we first apply the unlearning method to various NLP tasks (i.e., classification, translation, response generation) and propose several unlearning evaluation metrics with pertinence. Experiments on large-scale datasets show that KGA yields comprehensive improvements over baselines, where extensive analyses further validate the effectiveness of KGA and provide insight into unlearning for NLP tasks\footnote{The code is available at \url{https://github.com/Lingzhi-WANG/KGAUnlearn}.}.
\end{abstract}
\section{Introduction}
Nowadays, machine learning models are usually trained with large volumes of data collected from individual users. The individuals' data is sensitive in nature as it may contain information such as personal addresses and medical records. Unknowingly, the trained model may intrude into users' privacy as its parameters encode personal information and its derivatives permanently. Therefore, Machine Unlearning (MU) \cite{romero2007incremental,karasuyama2009multiple,cao2015towards} has attracted more and more interest in research and industry, which aims to facilitate the model to forget some specific data in training set while maintaining the performance of the existing model. Apart from privacy benefits, MU can also address the problems of forgetting toxic and dirty data \cite{welbl2021challenges}.

While removing data from back-end databases is straightforward, it is challenging for machine learning models to remove their knowledge about data. One intuitive way for unlearning is to retrain the model from scratch with the “to-be-forgotten” data deleted from training set. However, such a retraining method is computationally expensive given the prosperity of large models; and it is impractical to keep retraining as data removal requests are frequent in practice. Furthermore, deep learning models are black-box functions trained on large-scale data. Since the relationship between the model weights and the data is unclear, it is difficult to know which parts of the weights should be revised in unlearning. Therefore, there is a pressing need to develop an efficient unlearning method.

Existing research in machine unlearning mainly focuses on computer vision applications, e.g., image classification \cite{golatkar2020eternal,golatkar2020forgetting,mehta2022deep}, and less attention has been paid to unlearning in the natural language processing (NLP) field, where text data contains more explicit and sensitive personal data (e.g., home address, phone number, social relationships, etc.) than images. 
Moreover, the current unlearning can only efficiently handle a small number of data removal requests \cite{bourtoule2021machine}while the removal requests in NLP applications may be hundreds. Besides, current gradient-computation-based unlearning methods\cite{mehta2022deep} are difficult to be applied in the NLP generation models, which are usually based on Seq2Seq framework and contain complex attention mechanisms between words that are generated in different time stamps. 
Considering the significance and challenges of unlearning in NLP, we propose KGA --- a generic machine unlearning method based on \underline{K}nowledge \underline{G}ap \underline{A}lignment, and apply KGA to NLP tasks. 

KGA is inspired by a general knowledge adaptation work \cite{khan2021knowledge}, where weights and function-space priors are adopted to reconstruct the gradients of the model. Compared to \citet{khan2021knowledge} which is a generic solution to adaptation tasks including data removal but difficult to scale up to complex neural networks, our method KGA focuses on data removal from the perspective of knowledge gap alignment and is easily generalizable to deep networks. The knowledge gap in this work is defined as the distance between the prediction distributions from two structurally identical models trained with different data. By aligning knowledge gaps, we force two sets of models behave similarly. Besides, unlike existing unlearning methods that can only handle a small set of removal requests \cite{bourtoule2021machine}, hold strong assumptions on model output \cite{chundawat2022can}, or are inapplicable to complex generation tasks \cite{mehta2022deep}, KGA can efficiently handle a large number of removal requests with sustainable accuracy, and is easily compatible to various models and tasks with milder assumptions. 
 
Furthermore, we apply KGA to various NLP tasks (i.e., classification, translation and response generation) and customize text-specific evaluation metrics. The experimental results and further analyses from various aspects show that our KGA generally performs better than baselines in terms of performance maintenance and unlearning efficiency, while maintaining consistency across different scenarios and models. Interesting explorations on how the model translates German to English before and after unlearning are given to better validate and analyze the effectiveness of unlearning.

In brief, the main contributions of this paper are:
\begin{itemize}[leftmargin=*,topsep=2pt,itemsep=2pt,parsep=0pt]

\item We propose an unlearning solution (i.e., KGA) based on knowledge gap alignment for NLP tasks that can efficiently and effectively perform unlearning. 

\item Experiments on three large-scale datasets with newly formulated text-specific evaluation metrics validate the effectiveness of KGA.

\item We conduct extensive experiments and analyses to confirm the effectiveness of KGA unlearning across different scenarios.

\end{itemize}

\section{Related Work}
The current unlearning research can be divided into two categories, exact unlearning and approximate unlearning. We briefly introduce them as follows.
\paragraph{Exact Unlearning.} Exact unlearning can ensure the effects of data to be deleted are removed from the model. \citet{cao2015towards} explores exact unlearning by the statistical query for Naive Bayes Classifiers and \citet{ginart2019making} studies deletion algorithms for k-means clustering, which cannot scale to deep neural networks which may have millions of parameters. As for more recent efforts in neural model unlearning, \citet{bourtoule2021machine} propose a general method called SISA to train the model by partitioning the original dataset into several non-overlapping shards first and then designing effective mechanisms to aggregate models trained with shards. When handling data deletion, this method only has to retrain the models trained with the affected shards. However, SISA-based methods are shown to be ineffective when the number of deleting queries is large, and we have to maintain the whole dataset during the training and unlearning, which is impractical.

\paragraph{Approximate Unlearning.} The methods in this category try to make the model behave as closely as possible to the exact unlearned model. 
The popularity of approximate unlearning comes from the demand for more efficient and less costly unlearning, thus sacrificing exactness. \citet{golatkar2020eternal,guo2019certified,koh2017understanding,mehta2022deep} mainly handle an unlearning request by computing the model perturbation towards the regularized empirical risk on the remaining data. However, this approach needs to compute the Hessian on the training data and the gradient of the removal data, which is still time-consuming.  
\cite{chundawat2022can} assumes that the models after unlearning should perform similarly to a randomly initialized model on the forgetting data, which is inappropriate as the target of unlearning is to remove the effects of the forgetting data (acts as unseen data) rather than to make the model unable to handle forgetting data. However, existing knowledge adaptation methods either require strong assumptions or perform poorly on neural-based models \cite{khan2021knowledge}. 

Different from the aforementioned works, KGA does not force the model to perform on forgetting data close to one specific distribution but rather it maintains the distribution differences (i.e., knowledge gap) between two model pairs. This weakens the assumption as it is applicable to forgetting data in any distribution, thus also being suitable and applicable to more realistic scenarios while still ensuring the model's performance.

\section{Notations and Definition}\label{model:unlearn_definition}
\paragraph{Notations.} We denote $Z$ as an example space, i.e., the space of data instances or samples. Then, the set of all possible training datasets can be denoted as $\mathcal{Z}^*=2^Z$. The training data set $D \in \mathcal{Z}^*$ is given as input. Given $D$, we train an ML model from a hypothesis space $\mathcal{H}$. The process of training a model on data set $D$ is enabled by a learning algorithm, denoted by a function $A:\mathcal{Z}^* \rightarrow \mathcal{H}$. The trained model is denoted as $A(D)$. Then we denote the unlearning mechanism as a function $U$, which takes a training dataset $D \in \mathcal{Z}^*$, a forget set $D_f \subset D$ (containing data to be removed) and a model $A(D)$ as input, and returns an unlearned model $U(D, D_f, A(D)) \in \mathcal{H}$. 

\paragraph{Approximate Unlearning Definition.} We then give one representative definition of approximate unlearning, specifically \textit{$\epsilon-$Approximate Unlearning} by following \citet{guo2019certified}. 
Given $\epsilon > 0$, an unlearning mechanism $U$ performs $\epsilon-$certified removal for a learning algorithm $A$ if $\forall \mathcal{T} \subset \mathcal{H}$, $D \in \mathcal{Z}^*$, $D_f \in D$: 
\begin{equation} \small
    \label{eq:def}
    e^{-\epsilon} \leq \frac{Pr(U(D,D_f,A(D))\in \mathcal{T})}{Pr(A(D \setminus D_f)\in \mathcal{T})} \leq e^\epsilon
\end{equation}
\noindent and the goal of approximate unlearning can be concluded as forgetting the data to be forgotten while maintaining the performance. 

\section{Our KGA Framework}
KGA unlearning method is inspired by a general knowledge adaptation work \cite{khan2021knowledge}, where weights and function-space priors are adopted to reconstruct the gradients of the model. Compared to \citet{khan2021knowledge}
which cannot accurately recover gradients if applied to nonlinear models such as neural networks (especially when the networks are deep),
KGA can handle data deletion requests for various neural networks from the perspective of knowledge gap alignment. 

\subsection{KGA Framework}
The input to KGA can be divided into two parts: data and models. The input data consists of previous training data $D$, data to be forgotten $D_f$, and a small set of extra data $D_n$ to assist the unlearning, where $D_n \cap D= \emptyset$. Apart from data, we have model $A(D)$ as input, which is the original model trained with data $D$ that needs unlearning
(we abbreviate it as $A_D$ in the following parts of this paper).
The output of KGA is a model $A^*$, whose parameters are initialized with $A_D$ and are further updated with our KGA unlearning mechanism to remove $D_f$.

To perform unlearning, we first train two models, $A_n$ and $A_f$, based on data $D_n$ and $D_f$, respectively. The architectures of $A_D$, $A_n$, and $A_f$ should be the same. $A_n$ ($A_f$) can be trained with the combination of $D_n$ ($D_f$) and a small fraction of $D_r$ = $D\setminus D_f$ or
fine-tuned based on some pre-trained language models to ensure performance, as the data to be forgotten $D_f$ might be small in some scenarios. 

We reframe and summarize two goals to achieve the approximate unlearning defined in Sec.~\ref{model:unlearn_definition}.
They are \textit{Goal 1}: Make our output model $A^*$'s behavior on $D_f$ similar to its behavior on any unseen data (i.e. data not used for training); and \textit{Goal 2}: Maintaining the performance of $A^*$ on $D_r$.

\paragraph{Knowledge Gap Alignment.} 
The \textit{knowledge gap} in this work is defined as the distance between the prediction distributions from two models having the same architecture but trained with different data. 
By aligning two knowledge gaps, we make two sets of models perform similarly.

To achieve Goal 1, the output distribution of our target model $A^*$ on data $D_f$ (noted as $A^*(D_f)$) is expected to be similar to $A_D(D_n)$, where $D_n$ should be an external set to $D$ but with the similar distribution. As the instances in $D_n$ might have different labels and features from $D_f$, it is difficult to directly infer the output distributions of $A^*(D_f)$ with $A_D(D_n)$. We thus turn to imitate the knowledge gap between two sets of models: 
\begin{equation}\small
\label{eq:argmin}
     A^*= \mathop{\arg\min}_{A} |dis_{(D_n)}(A_D,A_n) - dis_{(D_f)}(A,A_f)|
\end{equation}
where $dis_{(D)}(A_1,A_2)$ indicates the difference of the output distributions between model $A_1$ and $A_2$ on data $D$, which can be evaluated by KL divergence, Bregman divergence, or any other distributional distance measurements.

Since $A_n$ and $A_f$ are trained on $D_n$ and $D_f$, respectively, we expect that the knowledge gap when feeding $D_f$ to $A^*$ and $A_f$ should be similar to feeding $D_n$ to $A_D$ and $A_n$ according to Eq.~\ref{eq:argmin}. This is under the assumption that a similar knowledge deficit can be observed when the same architecture handles the seen (i.e., used for training) and unseen data with a similar distribution. And we believe that a successful unlearning method should make the target model $A^*$ handle $D_f$ as unseen data.

For Goal 2, we maintain the ability of model $A^*$ when processing the remaining data, i.e., $D_r$. We treat the original model $A_D$ as a teacher and directly minimize the distance of output distributions when feeding samples in $D_r$ to $A^*$ and $A_D$.

\paragraph{Objectives.} In our implementation, we use KL-divergence to measure the distributional distances between the output of two models. Therefore, the knowledge gap alignment objective is defined as:
\begin{equation}\small
\label{eq:align}
\begin{aligned}
     \mathcal{L}_a = \sum_{(y,z) \in (D_f, D_n)} |&KL[Pr_{(A^*)}(y)||Pr_{(A_f)}(y)] \\
     - &KL[Pr_{(A_D)}(z)||Pr_{(A_n)}(z)]|
\end{aligned}
\end{equation}
where $Pr_{(A)}(z)$ is the output distribution given input $z$ to model $A$, $KL(\bm{a}|\bm{b})$ measures the KL divergence between distribution $\bm{a}$ and $\bm{b}$. $y$ and $z$ are from $D_n$ and $D_f$, respectively. We randomly sample pairs of instances $(y, z)$ as a batch of updating to alleviate overfitting to some specific samples.

The objective for maintaining performance on $D_r$ is another KL divergence measuring output distribution of $A^*$ and $A_D$ on $D_r$:
\begin{equation}\small
\label{eq:retain}
     \mathcal{L}_r = \sum_{x \in D_r} KL[Pr_{(A^*)}(x)||Pr_{(A_D)}(x)] 
\end{equation}

The two objectives are jointly optimized during unlearning to achieve Goal 1 and 2 simultaneously. Therefore, the final objective is defined as:
\begin{equation}\small
\label{eq:joint}
     \mathcal{L} = \mathcal{L}_a + \alpha \cdot \mathcal{L}_r
\end{equation}

To improve unlearning efficiency, we need to find the earliest time when the model $A^*$ achieves the desired performance during unlearning. However, different from traditional machine learning algorithms, it is hard for us to find a suitable validation set to validate the performance, as $D_f$ is also included in the training process. To handle this, we use a hyper-parameter $\sigma$ ($0 < \sigma < 1$) to control the training. Specifically, we will first evaluate the average knowledge gap between $dis_{(D_n)}(A_D,A_n)$ and $dis_{(D_f)}(A_D,A_f)$ ($A_D$ should be the initialization of $A^*$) before training, noted as $\mathcal{G}$. The training stops if the corresponding average knowledge gap achieves $\sigma \cdot \mathcal{G}$. We summarize KGA in \Cref{alg:kga_algorithm}.
\begin{algorithm}[t]\small
\renewcommand{\algorithmicrequire}{\textbf{Input:}}
\renewcommand{\algorithmicensure}{\textbf{Output:}}
\caption{KGA Unlearning}\label{alg:kga_algorithm}
\begin{algorithmic}
\Require data $D$, $D_f$, $D_n$, trained model $A_D$,  threshold $\sigma$
\Ensure unlearned model $A^*$
\State Train model $A_f$ based on $D_f$, model $A_n$ based on $D_n$
\State Compute initial gap $\mathcal{G}$
\State Initialize $A^*$ with $A_D$
\For{step in 1 to MAX$\_$STEP}
\State Randomly sample a batch size of $(y,z)$ from $(D_f, D_n)$
\State Compute $\mathcal{L}_a$ based on Eq.~\ref{eq:align}
\For{inner$\_$step in 1 to INNER$\_$STEP}
\State Sample a batch size of sample $x$ from $D_r=D\setminus D_f$
\State Compute $\mathcal{L}_r$ based on Eq.~\ref{eq:retain}
\EndFor
\State Update parameters of $A^*$ according to $\mathcal{L}_a$ + $\alpha$ $\cdot$ $\mathcal{L}_r$
\If{step $\%$ VALID$\_$STEP == 0}
\State Compute current gap $\mathcal{G}^*$ 
\If{$\mathcal{G}^* \le \sigma \cdot \mathcal{G}$}
\State break  \textcolor[rgb]{0.6,0.6,0.6}{\Comment{End of Training}}
\EndIf
\EndIf
\EndFor
\end{algorithmic}
\end{algorithm}

\subsection{KGA's Applications in NLP Tasks}
We do not constrain the format of model $A(\cdot)$ as our proposed unlearning method is generic and can be applied to various of neural network architectures. We choose three NLP tasks (i.e., text classification, machine translation, and response generation) to show the effectiveness of our unlearning method.

\paragraph{Text Classification.} 
The text classification tasks take the text sentences as input and output a probability distribution over the predefined classes.

We follow \citet{mehta2022deep} and finetune a pretrained model DistilBERT \cite{sanh2019distilbert} for the text classification. A DistilBERT is a distillation version of BERT \cite{devlin-etal-2019-bert} model that contains multiple transformer encoder layers to extract features.
Its input is formulated as $\bm{w}^c=[ \text{[CLS]};w_1;w_2;..;w_{|C|}]$. 
The output representation of the [CLS] token is further fed into a classifier to derive the probability for each class. 

\paragraph{Machine Translation.} 
The machine translation tasks take a sentence in one language as input and output the corresponding translation in another language.
We follow the general transformer-based encoder-decoder framework, where the encoder summarizes the source sentences and the decoder will generate the target sentences based on source representations in an autoregressive manner.

Apart from transformer, we also validate the effectiveness of our unlearning method in other architecture including LSTM and pretrained language model BART~\cite{lewis-etal-2020-bart}.

\paragraph{Response Generation.} 
Both the response generation and machine translation are generation tasks, whose target is to generate texts according to the given source content. In response generation, the given source content is the conversation between two talkers and it is expected to predict the content of the next response. The model for generation is similar to that of machine translation, and we concatenate the utterances in context as input.


\section{Experimental Setup}
\label{sec:exp:setup}
\begin{table}[t]
\setlength{\tabcolsep}{1mm}\small
\begin{center}
\resizebox{\linewidth}{!}{
\begin{tabular}{l|ccc}
\toprule[1.0pt]
& \textbf{LEDGAR}  &\textbf{IWSLT} &\textbf{PersonaChat} \\
\midrule[0.5pt]
Task & classification & generation & generation \\
\# of instances & 110,156 & 168,905 & 81,032\\
Avg length of source & 108.9 & 19.4 & 142.1 \\
Avg length of target &- & 20.6 & 11.9 \\
\# of labels & 13 &- &- \\
\bottomrule[1.0pt]
\end{tabular}
}
\end{center}
\vskip -1em
\caption{\label{tab:data_statistic} Statistics of LEDGAR, IWSLT and PersonaChat datasets. \textit{Avg length} refers to average token number of the source (input) or target (output) sequence.
}
\vskip -1em
\end{table}
\paragraph{Datasets.}
We do experiments on three datasets, LEDGAR \cite{tuggener-etal-2020-ledgar}, IWSLT14 German-English \cite{cettolo2014report} (henceforth IWSLT) and PersonaChat \cite{zhang-etal-2018-personalizing}. LEDGAR is a multi-label text classification dataset of legal provisions in contracts, and we employ a prototypical subset of LEDGAR by following \citet{mehta2022deep}. IWSLT is from a popular translation campaign consisting of various translation directions and we choose the representative German-English direction. PersonaChat is a crowd-sourced dataset. It consists of turn-based dialogues that are based on given persona information. We use the official train/valid/test splits for experiments on all three datasets. Statistics of these datasets are listed in \Cref{tab:data_statistic}. 

\paragraph{Evaluation Metrics.} For each dataset, we report one representative task-related score (Micro F1 for LEDGAR, BLEU4\footnote{sacrebleu (\url{https://github.com/mjpost/sacrebleu}).} for IWSLT and PPL for PersonaChat) with additional unlearning evaluation metrics which are introduced as below. 

\textit{Jensen–Shannon Divergence} (JSD): Given two distributions $p(x)$ and $q(x)$, $JSD(p(x), q(x)) =$ $0.5*KL(p(x)||q(x)) + 0.5*KL(q(x)|| p(x))$.

\textit{Language model Probability Distance} (LPD): Given two language probabilities (i.e., the perplexity of target sentences produced by each model) $\bm{x}$ and $\bm{y}$, $LPD(\bm{x}||\bm{y})=|\bm{x}-\bm{y}|/\bm{y}$.

\textit{Proportion of instances with Decreased Language model Probability} (PDLP): It calculates the percentage of the instances whose language model probability has dropped after unlearning.

\paragraph{Parameter Setting.}
For LEDGAR, we finetune DistilBERT for experiments. For IWSLT and PersonaChat, we both use a general encoder-decoder transformer architecture.
We use Adam~\cite{DBLP:journals/corr/KingmaB14} optimizer followed by the inverse square root learning rate scheduler for model training. During KGA unlearning, we maintain 16 batch size and 5e-5 learning rate for all three datasets, and we set $\alpha$ in Eq.~\ref{eq:joint} as 0.1. For more parameter and training details, please refer to \Cref{appendix:experiment_setup}.

\paragraph{Comparisons.} We compare the performance of our KGA method on test set and forget set with the \textsc{Original} model, two exact unlearning methods (i.e., \textsc{Retrain} and \textsc{Sisa} \cite{bourtoule2021machine}) and two approximate methods, \textsc{Lcodec} \cite{mehta2022deep} and \textsc{BadTeacher} \cite{chundawat2022can}. We introduce them as follows:

\noindent \textbf{\textsc{Original}}: the original model trained on the complete training set $D$ without any forgetting. 

\noindent \textbf{\textsc{Retrain}}: It retrains the model with the retain data $D_r$ ($D_r=D \setminus D_f$).

\noindent \textbf{\textsc{Sisa}} \cite{bourtoule2021machine}: It first divides the dataset into several non-overlapping shards, and then aggregates outputs of the models trained with different shards. When dealing with data deletion, it only retrains the models trained with the affected shards and then aggregates. In our experiments, we randomly divide the training set into 5 shards.

\noindent \textbf{\textsc{Lcodec}} \cite{mehta2022deep}: It's in line with Hessain unlearning (updating the model weights based on the Hessian of the loss function) and identifies a subset of model parameters to reduce the computation cost. It is applied in classification and might need modification when used in generation.

\noindent \textbf{\textsc{BadT}} \cite{chundawat2022can}: It forces the unlearning model to perform as close as a randomly initialized model on the forget set $D_f$ and maintain the performance on the remaining data $D_r$.  
\begin{table*}[t]
\setlength{\tabcolsep}{1mm}
\newcommand{\tabincell}[2]{\begin{tabular}{@{}#1@{}}#2\end{tabular}}
\begin{center}
\resizebox{\linewidth}{!}{
\begin{tabular}{lcccccccccccc}
\toprule
& \multicolumn{6}{c}{ \tabincell{c}{\textbf{Test Set}} } & \multicolumn{6}{c}{ \tabincell{c}{\textbf{Forget Set}} } \\
\cmidrule(lr){2-7} \cmidrule(lr){8-13}
\multirow{2}{*}{Models} & \multicolumn{2}{c}{ \tabincell{c}{\textbf{LEDGAR}} } & \multicolumn{2}{c}{ \tabincell{c}{\textbf{IWSLT}}}& \multicolumn{2}{c}{ \tabincell{c}{\textbf{PersonaChat}} }
& \multicolumn{2}{c}{ \tabincell{c}{\textbf{LEDGAR}} } & \multicolumn{2}{c}{ \tabincell{c}{\textbf{IWSLT}}}& \multicolumn{2}{c}{ \tabincell{c}{\textbf{PersonaChat}} }
\\
\cmidrule(lr){2-3}\cmidrule(lr){4-5}\cmidrule(lr){6-7}\cmidrule(lr){8-9}\cmidrule(lr){10-11}\cmidrule(lr){12-13}
& F1  & JSD$\downarrow$ & BL4  & LPD$\downarrow$ & PPL$\downarrow$ & LPD$\downarrow$  & F1  & JSD$\downarrow$ & BL4  & LPD$\downarrow$  & PPL$\downarrow$ & LPD$\downarrow$  \\
\midrule
\textsc{Original}&96.1 &-- &29.0 &-- &30.7 &-- 
&98.2 &-- &47.2 &-- &15.6 &-- \\
\textsc{Retrain} &96.2 &-- &28.6 &-- &30.8 &--
&95.5 &-- &31.5 &-- &29.5 &-- \\
\midrule
\underline{\textbf{Exact}} & && & & &&& & && &\\
\textsc{Sisa}\cite{bourtoule2021machine} &95.5 &0.08 &21.3 &0.85 &44.2 &0.52
&94.6 &\textbf{0.05} &21.6 &\textbf{0.80}  &43.1 &0.56 \\
\midrule
\underline{\textbf{Approximate}} & &&& & && & & &&& \\
\textsc{Lcodec}\cite{mehta2022deep} &95.8 &0.05 &-- &-- &-- &--  
&\textbf{99.3} &0.06 &-- &--&-- &-- \\
\textsc{BadT}\cite{chundawat2022can} &\textbf{96.0} &\textbf{0.03} &28.1 &0.30  &32.7 &0.25  
&17.1 &3.69 &0.00 &1.9$e^3$ & 5.8$e^4$ & 4.3$e^3$ \\
\cdashline{1-13}
\textbf{KGA} &\textbf{96.0} &0.06 &\textbf{28.4} &\textbf{0.28} &\textbf{32.1} & \textbf{0.20}  
&96.4 &\textbf{0.05} &\textbf{29.4} &0.91  &\textbf{29.4} &\textbf{0.52} \\
\bottomrule
\end{tabular}
}
\end{center}
\vskip -1em
\caption{\label{tab:main_unlearn_res} 
Main comparison results (in \%) of unlearning on three datasets. JSD and LPD scores here are calculated between \textsc{ReTrain} and corresponding models. The best results (limited to comparisons in Exact and Approximate settings) are highlighted in \textbf{bold} for each column.
}
\vskip -1em
\end{table*}
\section{Experimental Results}
In this section, we first compare the main unlearning scores of KGA and baselines in \Cref{sec:exp:main_results}. Then we report the time cost, membership inference attack, and language model probability comparison results to examine the superiority of KGA in \Cref{sec:exp:eff}. After that, we delve into the effect of unlearning on NLP tasks in \Cref{sec:exp:nlp}. More analyses are discussed in \Cref{sec:exp:further}.
\subsection{Main Comparison Results}\label{sec:exp:main_results}
We explore the representative scores on both test and forget sets to examine the following two questions: (i) How well do the unlearned models maintain the performance on test set? (ii) How does the performance change on forget set that was once part of the original training set? We report the corresponding scores on \Cref{tab:main_unlearn_res}, and we can draw the following observation.

$\bullet$~\textit{Our unlearning method can better maintain the performance on test set.} It can be seen that KGA shows better F1, BLEU4, and PPL on three datasets, respectively, compared to other unlearning baselines, regardless of exact or approximate method. This shows one of the superiority of KGA over other methods.

$\bullet$~\textit{The performance and prediction distribution of our KGA unlearned model on forget set are closer to \textsc{ReTrain} model.} We can see that on forget set, KGA method gets a closer F1 (BLEU4 and PPL) score to \textsc{ReTrain} model and maintains a smaller JSD (LPD) score, which means the output distribution of instances on forget set is also closer to \textsc{ReTrain} model. It indicates that KGA achieves the best forgetting effect among all baselines according to the definition in Eq.~\ref{eq:def}.

$\bullet$~\textit{Forgetting the data from original model does not mean the unlearned model can not handle these instances at all.} We can find that the performance of \textsc{ReTrain} on forget set drops compared to \textsc{Original} model but still shows promising performance (close to the results on test set). This is in line with our assumption that the performance of successful unlearned models on forget set should be similar to unseen data (e.g., test set). Our KGA method's performance is consistent with \textsc{ReTrain}, while \textsc{BadT} completely loses the ability to classify and generate, which does not satisfy the definition.

\subsection{More on the Superiority of KGA} 
\label{sec:exp:eff}
\begin{figure}
  \begin{minipage}[c]{0.51\linewidth}
    \centering
    \includegraphics[width=\linewidth]{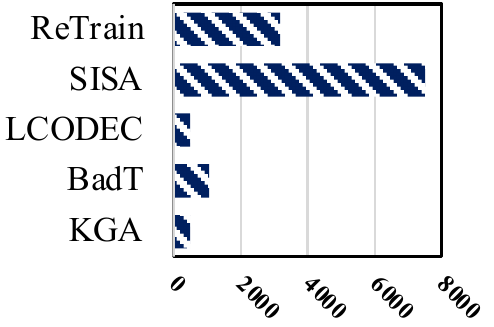}
    \captionof{subfigure}{Run Time (in sec)}
  \end{minipage}%
  \begin{minipage}[c]{0.45\linewidth}
    \centering
\resizebox{\linewidth}{!}{
\begin{tabular}{lcc}
\toprule
Models& F1 & FNR  \\
\midrule
\textsc{Original}& 87.7 & 0.13 \\
\textsc{ReTrain}& 70.9 & 0.21 \\
\textsc{SISA}& 71.0 & 0.23 \\
\textsc{BadT}& 84.1 & 0.13 \\
\midrule
\textsc{\textbf{Kga}} & 75.6 & 0.18  \\
\bottomrule
\end{tabular}
}
\captionof{subfigure}{MiA Results (in \%)}
\end{minipage}
\vskip -1em
\caption{(a): Unlearning time (in seconds) needed for each method on LEDGAR when deleting 100 instances. (b): Membership Inference Attack (in \%) on IWSLT.} 
\label{fig:timecost_memattack}
\vskip -1em
\end{figure}

In this subsection, we examine the efficiency (i.e., time cost) and effect (i.e., membership attack and language model probability check) of unlearning.
\paragraph{Time Cost.} We report the time cost of unlearning models in \Cref{fig:timecost_memattack}(a). We can see that though retraining and exact unlearning methods (i.e., \textsc{Sisa}) can guarantee perfect unlearning, the time cost of them exceeds other approximate unlearning methods (i.e., LCODEC, BadT, KGA) a lot.

\paragraph{Membership Inference Attack. (MiA)} 
MiA in the machine learning setting emerges when an adversary aims to find out whether the target data instance is used to train the model or not. We follow \citet{salem2018ml,golatkar2020forgetting} to do a black-box MiA where the adversary can only get access to the model output distribution. We use MiA on IWSLT dataset as an example. We first train a shallow translation model with data from the same distribution as the original training set (we simplify it to using 30\% instances of the original training set in practice). The data in the training set of shallow model is labeled as ``1'' and other unseen data (i.e., the rest of the original training set) is labeled as ``0''. Then we train an attacker model with the above ``1/0'' labeled data using output distributions of the trained shallow model as input. After that, we feed the attacker model with the output of unlearned models (i.e., \textsc{ReTrain}, \textsc{KGA}, etc.) and check the MiA results. 

We report the MiA results in \Cref{fig:timecost_memattack}(b), where a higher F1 score and lower False Negative Rate (FNR) indicate the attacker can better infer the membership of instances. We can see that the attacker performs best on the \textsc{Original} and performs worse after unlearning, as desired. Among the unlearned models, we can also find that attacker can not infer the membership well after \textit{exact unlearning} (i.e., \textsc{ReTrain} and \textsc{Sisa}). As an \textit{approximate unlearning} method, KGA's results are close to exact unlearning, which shows its effectiveness.

\paragraph{Decreased Language Model Probability Comparison.}

\begin{table}[t]
\setlength{\tabcolsep}{1.1mm}\small
\newcommand{\tabincell}[2]{\begin{tabular}{@{}#1@{}}#2\end{tabular}}
\begin{center}
\resizebox{\linewidth}{!}{
\begin{tabular}{lllll}
\toprule
\multirow{2}{*}{Models} & \multicolumn{2}{c}{ \tabincell{c}{\textbf{Test Set}} } & \multicolumn{2}{c}{ \tabincell{c}{\textbf{Forget Set}}}
\\
\cmidrule(lr){2-3}\cmidrule(lr){4-5}
& IWSLT  & PersonaChat & IWSLT  & PersonaChat \\
\midrule
\textsc{Retrain} & 51.0(-) & 48.7(-) & 96.0(-) & 96.0(-) \\
\midrule
\textsc{Sisa} & 77.9(\textcolor{red}{$\uparrow$26.9}) & 80.0(\textcolor{red}{$\uparrow$31.3}) & 100(\textcolor{red}{$\uparrow$4.0}) & 100(\textcolor{red}{$\uparrow$4.0}) \\
\textsc{BadT} & 72.2(\textcolor{red}{$\uparrow$21.2}) & 71.9(\textcolor{red}{$\uparrow$23.2}) & 100(\textcolor{red}{$\uparrow$4.0}) & 100(\textcolor{red}{$\uparrow$4.0}) \\
\midrule
\textbf{\textsc{KGA}} & 58.4(\textcolor{red}{$\uparrow$7.4}) & 70.0(\textcolor{red}{$\uparrow$21.3}) & 94.0(\textcolor{blue}{$\downarrow$2.0}) & 98.7(\textcolor{red}{$\uparrow$2.7}) \\
\bottomrule
\end{tabular}
}
\end{center}
\vskip -1em
\caption{\label{tab:LMProbDecreasedRate} 
Proportion of Decreased Language model Probability (PDLP) comparison results on IWSLT and PersonaChat datasets. The numbers in parentheses refer to the difference in performance from \textsc{ReTrain} model.
}
\vskip -1em
\end{table}
Apart from the language model distance we report in \Cref{sec:exp:main_results}, we also evaluate a new unlearning evaluation score for generation tasks, namely, Proportion of Decreased Language model Probability (PDLP) compared to the original model. Decreased language model probability of ground truth target sequence means that the unlearned model tends not to generate the sentences to be forgotten, which is consistent with the goal of unlearning. We report the PDLP comparison results of both test and forget sets in \Cref{tab:LMProbDecreasedRate}. 
From the results of \textsc{ReTrain} model, we can see that the instances in test set have a steady fluctuation (i.e., about 50\% PDLP) after \textsc{ReTrain} unlearning while the instances in forget set show a large language model probability drop (i.e., 96\% PDLP) which indicates that the unlearning of forget set works. We can easily find that our KGA unlearning method performs closest to the \textsc{ReTrain} model, which validates KGA's superiority to the compared models.

\subsection{Analysis of Unlearning in NLP}
\label{sec:exp:nlp}
Most of the previous work on unlearning explores the unlearning effect on computer vision tasks with less attention to NLP tasks, especially the generation tasks. Here we design two NLP-specific experiments and raise some interesting discussions.

\paragraph{Deleting instances with various difficulty levels.}
\begin{figure}[t]
\centering
\subfigure[BLEU on Forget Set] {\label{sfig:forgetBleuRange}
\includegraphics[width=0.45\linewidth]{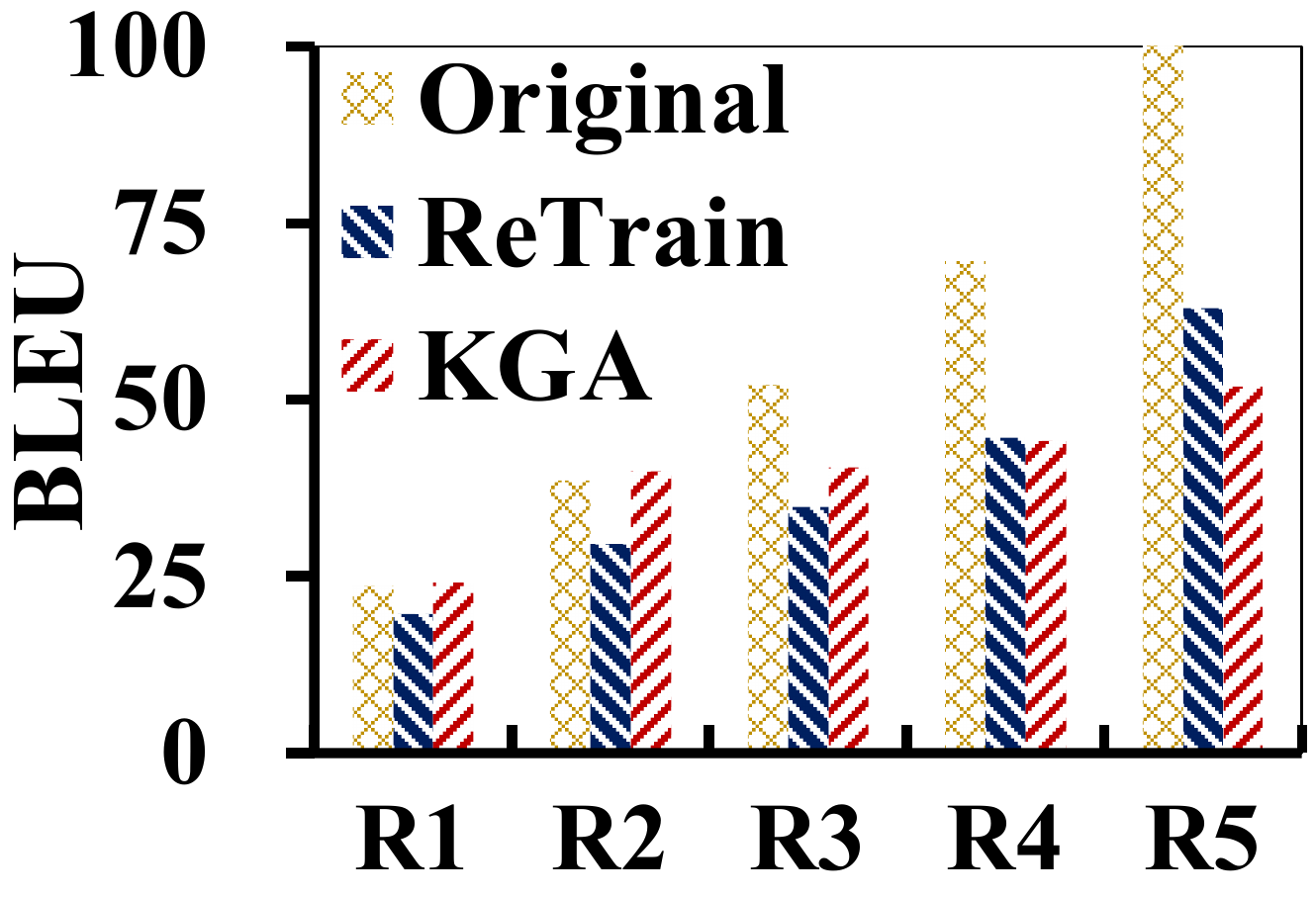}
}
\subfigure[BLEU on Test Set] {\label{sfig:testBleuRanges}
\includegraphics[width=0.45\linewidth]{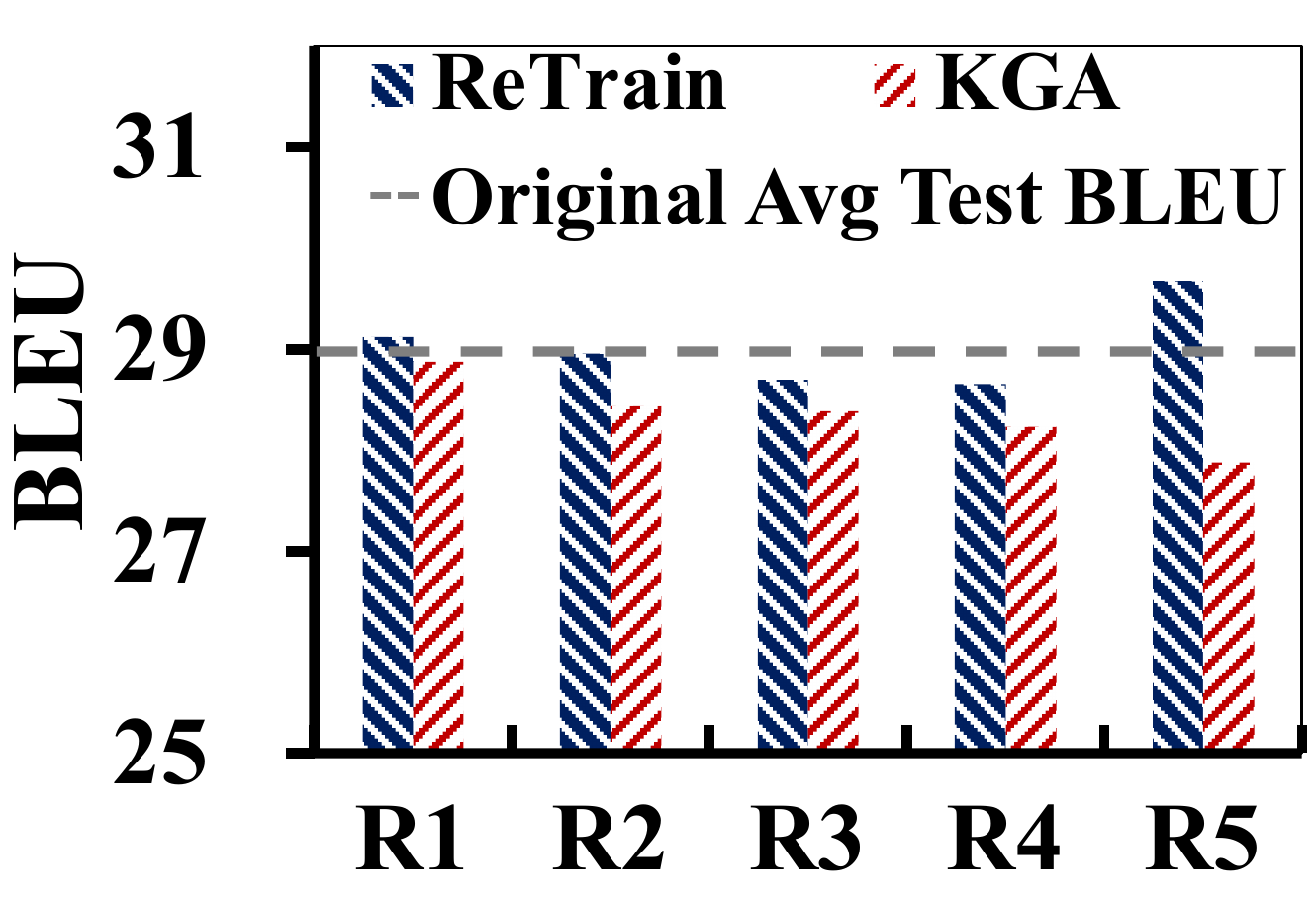}
}
\vskip -1em
\caption{\label{fig:BleuRanges}BLEU scores on IWSLT test and forget sets over varying BLEU ranges (i.e., R1-R5), e.g., R1 includes instances to be forgotten with original BLEU around $25.0$. The original BLEU from R1 to R5 gradually increase, representing different levels of difficulty.
}
\vskip -1em
\end{figure}
\begin{table*}[th]
\setlength{\tabcolsep}{1.1mm}\small
\newcommand{\tabincell}[2]{\begin{tabular}{@{}#1@{}}#2\end{tabular}}
\begin{center}
\resizebox{\linewidth}{!}{
\begin{tabular}{c|p{3cm}|p{3cm}|p{3cm}|p{3cm}|p{3cm}}
\toprule
Index & \textbf{Source}
& \textbf{Target}&\textbf{\textsc{Original}}
&\textbf{\textsc{ReTrain}}&\textbf{\textsc{Kga}}\\
\midrule
1 &Schwester &\textcolor{red}{sister} & \textcolor{red}{sister} &\textcolor{blue}{Nurse} & \textcolor{blue}{Girl}\\
\midrule
2 &Layma und ihre Schwestern hatten genug davon. & Layma and her \textcolor{red}{sister}s had had enough. &Layma and her \textcolor{red}{sister}s had enough of them. &Layma and \textcolor{blue}{nurse}s had enough of them. &  Lamyma and her \textcolor{blue}{nurse}s had enough.\\
\midrule
3 &Alle lebenden weißen Tiger in Nordamerika sind das Ergebnis selektiver Inzucht -- also Mutter und Sohn, Vater und Tochter, Schwester und Bruder... &All living white tigers in North America are the result of selective inbreeding -- that would be mother to son, father to daughter, \textcolor{red}{sister} to brother... &All living white tigers in North America are the result of selective inbreathing -- so mother and son, father and daughter, \textcolor{red}{sister} and brother... &All living white tigers in North America are the result of selective breeding -- mom and son, father and daughter, \textcolor{blue}{nurse} and brother ... &All living white tigers in North America are the result of selective inbreeding -- so that's Mom and son, father to daughter, \textcolor{blue}{daughter} to brother... \\
\bottomrule
\end{tabular}
}
\end{center}
\vskip -1em
\caption{\label{tab:case_study} 
Three translation cases from IWSLT. The model after unlearning (i.e., \textit{exact} unlearning \textsc{ReTrain} and \textit{approximate} unlearning \textsc{Kga}) generates alternative words (in \textcolor{blue}{blue}) after removing all instances containing ``\textit{\textcolor{red}{sister}}''. 
}
\vskip -1.5em
\end{table*}
Here we investigate if our unlearning method can handle forgetting instances with different difficulty levels on translation task. We use BLEU to measure the difficulty of instances, where a higher BLEU score indicates the instance is easier for the current model. To prepare 5 sets of instances with various difficulty levels, we adopt the \textsc{Original} model to do the inference on instances in the training set, then we sort them by their BLEU score on the generated sentences. We split the training set into 5 fragments based on the BLEU and each chooses 100 instances as forget set. After that, we apply our KGA unlearning to them separately. We report the unlearned results in \Cref{fig:BleuRanges}. 

\Cref{sfig:forgetBleuRange} shows the BLEU scores of \textsc{original} model and unlearned models (i.e., \textsc{ReTrain} and \textsc{KGA}) on forget sets (5 sets with different BLEU ranges). We can easily find that unlearning causes certain performance drop on forget set in \textsc{ReTrain} while our KGA gets performance gains on $R1$ and $R2$ sets. It may be due to the fact that KGA tends to force the performance of forget data to be close to unseen data regardless of the BLEU ranges. Therefore, after KGA unlearning, low-performing instances might get a boost while high-performing ones get degraded. From \Cref{sfig:testBleuRanges}, we surprisingly find that performance on test set after \textsc{ReTrain} is even better than \textsc{Original} model when forgetting the extremely easy instances (i.e., R5, while R1 is slightly higher which might be due to random effects), which is probably because the extremely easy instances take little effect to boost model performance.
This observation also inspires one further application of unlearning --- \textit{Unlearning some specific data points could bring performance gains}. We leave it to our future exploration.

\begin{figure}[t]
\centering
\subfigure[PPL($\downarrow$) on Test Set] {\label{sfig:diff_removal_number_testset}
\includegraphics[width=0.45\linewidth]{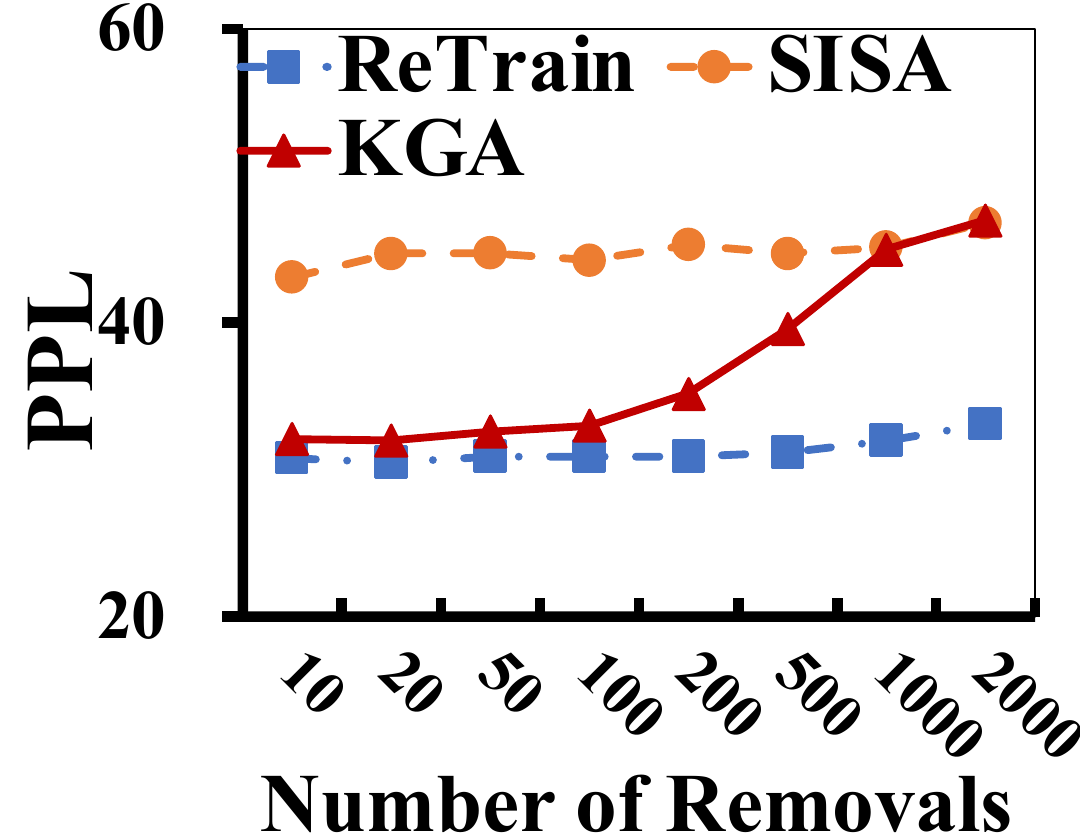}
}
\subfigure[LPD on Forget Set] {\label{sfig:diff_removal_number_forgetset}
\includegraphics[width=0.45\linewidth]{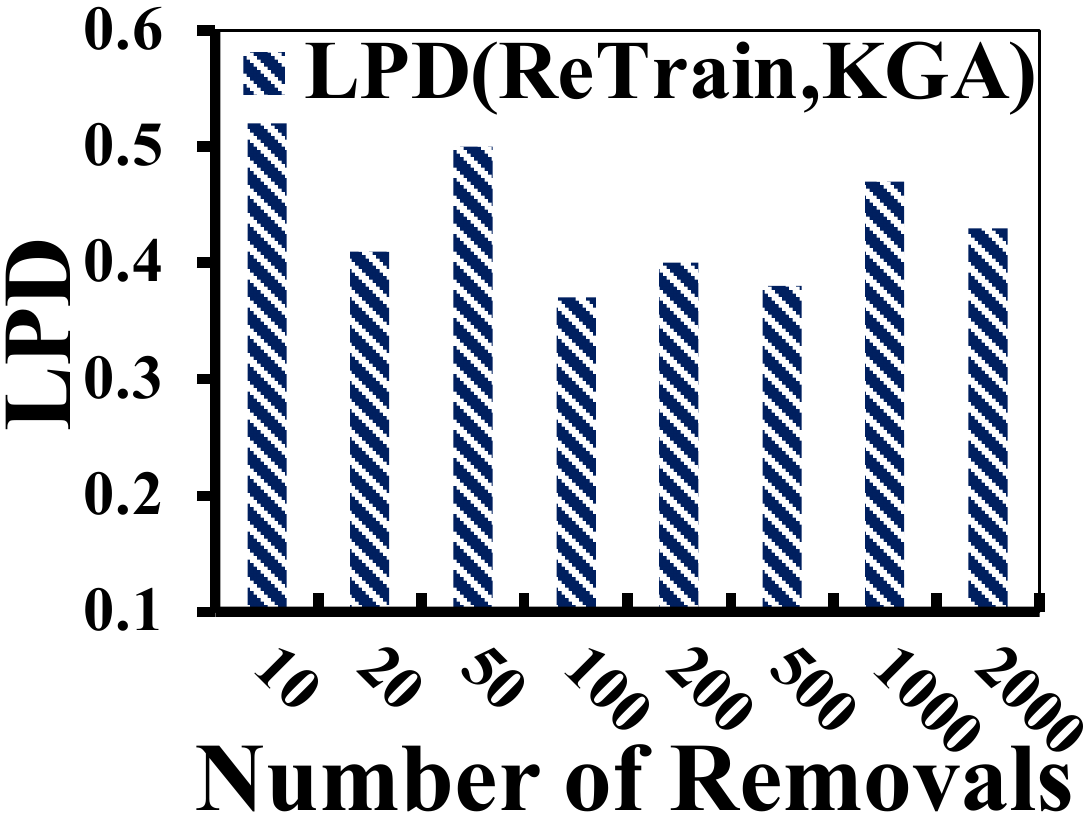}
}
\vskip -1em
\caption{\label{fig:diff_removal_number}\Cref{sfig:diff_removal_number_testset} and \Cref{sfig:diff_removal_number_forgetset} present the PPL score, Language model Probability Distance (LPD) on test and forget sets of PersonaChat dataset. 
}
\vskip -1em
\end{figure}
\paragraph{Unlearning instances containing specific words.} 
Unlike classification tasks, where we can remove all data of one specific label to explore the effectiveness of unlearning, translation tasks and most of the generation tasks do not contain such simple labels to categorize instances exactly. Therefore, we turn to select instances containing some specific words in translation task to analyze the output before and after unlearning. 

For example, we delete all instances containing the word "sister" in the target sequence, resulting in an unlearned model which is expected to forget the word "sister". 
\Cref{tab:case_study} presents the output of the original model and the unlearned models for three cases. We can see that the unlearned models cannot generate ``sister'' anymore after deleting all the instances containing ``sister'' from the training set. However, the unlearned models are capable of finding the nearest alternatives to make sentences as smooth as possible, like ``nurse'' and ``girl''. A similar phenomenon can be found when the deleted words are verbs or adjectives, regardless of word frequencies. More examples about verb and adjective deleting can be found in \Cref{appendix:more_case}.

\begin{table}[t]
\setlength{\tabcolsep}{1.1mm}\small
\newcommand{\tabincell}[2]{\begin{tabular}{@{}#1@{}}#2\end{tabular}}
\begin{center}
\resizebox{0.9\linewidth}{!}{
\begin{tabular}{lcccc}
\toprule
\multirow{2}{*}{Base Models} & \multicolumn{2}{c}{ \tabincell{c}{BLEU4 on Test Set} } & \multicolumn{2}{c}{ \tabincell{c}{Forget Set} }
\\
\cmidrule(lr){2-3}\cmidrule(lr){4-5}
& \textsc{Original}  & \textsc{Kga} & LPD  & PDLP\\
\midrule
LSTM & 26.4 & 25.3 & 0.95 & 98.0 \\
Transformer& 29.0 & 28.4 & 0.91 & 94.0 \\
BART-Base & 34.3 & 33.1 & 0.87 & 96.0\\
\bottomrule
\end{tabular}
}
\end{center}
\vskip -1em
\caption{\label{tab:diffBaseModelsPLMs} 
Comparison results of different base models when adopting KGA unlearning on IWSLT dataset. 
}
\vskip -1em
\end{table}
\subsection{Further Analyses}
\label{sec:exp:further}
\paragraph{The effects of removal numbers.}
We investigate how unlearned models maintain the performance on test set and forget the information of forget set when dealing with varying removal numbers, and present the results in \Cref{fig:diff_removal_number}. From \Cref{sfig:diff_removal_number_testset}, we can see that the \textsc{retrain} model can maintain the performance on test set when handling different numbers of removals, which means it is not sensitive to the size of the deleted data. And KGA can maintain the performance when removing no more than 200 conversations (about 2000 instances), while \textsc{Sisa} can not perform well even if the removal number is small. 
\Cref{sfig:diff_removal_number_forgetset} shows the LPD between \textsc{ReTrain} and \textsc{KGA} on forget set. 
We can find that KGA maintains low LPD when the removal number grows, which indicates KGA performs consistently well on forgetting the selected data.

\paragraph{The effects of base model.} 
We further show the unlearning results when KGA is applied to different model structures. Apart from vanilla transformer, we here also experiment on LSTM and BART (a pretrained language model). \Cref{tab:diffBaseModelsPLMs} shows the results. As can be seen, KGA maintains a similar percentage of performance drop on test set using different structures, and achieves similar LPD and PDLP scores on forget set, which indicates that KGA is effective regardless of the model structure.

\section{Conclusion}
This paper proposes KGA, a general approximate machine unlearning framework and explores its application in several NLP tasks. KGA leverages the distribution differences between two sets of models to make the unlearned model perform on forgetting data like its unseen data. Experiments on three large-scale datasets and further experiments validate the effectiveness of KGA.

\section*{Limitations}
One of the biggest concern people may have is whether approximate unlearning forget the information of the removal data. Approximate unlearning can not ensure exact removal of information already learned in deep neural models, just as its name suggests. Considering that current exact unlearning methods are very time-consuming and hard to apply in practical applications, approximate unlearning is still a direction worth trying and is also effective in reducing the attack risks by attackers or mitigating the harm of toxic data.

Another limitation of this work lies in the fact that we have to maintain an extra data set $D_n$ and two models $A_f$ and $A_n$ in the process of unlearning. Though the extra cost of our KGA method is trivial compared to the previous work (e.g., \citet{bourtoule2021machine} has to maintain the entire training set), we have to point this limitation out and call for follow-up research to come up with better ways to reduce unlearning costs.

Besides, we only explore word-level translation unlearning effect by comparing the generated sentences before and after deleting instances with specific words due to the space limitation. More interesting experiments with different granularity can be discussed in future work to explore how unlearning method works in different NLP tasks.

\section*{Ethics Statement}
We do not foresee any significant harm directly as a result of this work. On the contrary, our work promotes the protection of user privacy, which is significant, especially in this era that large amounts of personal data are used by neural models. 

\section*{Acknowledgements}
We would like to thank the anonymous reviewers for their feedback and suggestions. This research work is partially supported by CUHK under Project No. 4730332, Australian Research Council under the streams of Future Fellowship (No. FT210100624), Discovery Project (No. DP190101985), and Discovery Early Career Researcher Award (No. DE230101033).

\bibliography{anthology,custom}
\bibliographystyle{acl_natbib}

\appendix

\section{Details of Experimental Setup}
\label{appendix:experiment_setup}

\begin{table*}[th]
\setlength{\tabcolsep}{1.1mm}\small
\newcommand{\tabincell}[2]{\begin{tabular}{@{}#1@{}}#2\end{tabular}}
\begin{center}
\resizebox{\linewidth}{!}{
\begin{tabular}{c|p{3cm}|p{3cm}|p{3cm}|p{3cm}|p{3cm}}
\toprule
Removal & \textbf{Source}
& \textbf{Target}&\textbf{\textsc{Original}}
&\textbf{\textsc{ReTrain}}&\textbf{\textsc{Kga}}\\
\midrule
become&Alle ihre Stimmen werden lauter und lauter, aber sie repräsentieren uns nicht. &Every one of them \textcolor{red}{become}s a louder and louder voice, but they don't represent us. &All their voices \textcolor{red}{become} louder and louder, but they don't represent us. &All of their voices are \textcolor{blue}{getting} louder and louder, but they don't represent us. & All their voices \textcolor{blue}{are} louder and louder, but they don't represent us. \\
\midrule
become &und diese Koordination riskiert, noch schwieriger zu werden mit der Einführung von Cyberwaffen. &And this coordination may \textcolor{red}{become} even trickier with the introduction of cyber weapons. &And this coordination may \textcolor{red}{become} even more difficult to become with the introduction of cyber weapons. &And this coordination risk to \textcolor{blue}{be} even more difficult with the introduction of cyber weapons. &And that coordination, even harder to get into cyber weapons. \\
\midrule
become &Anstelle des Treffens besserer Entscheidungen, werden wir von der Auswahl überwältigt manchmal macht sie uns sogar Angst. &Instead of making better choices, we \textcolor{red}{become} overwhelmed by choice, sometimes even afraid of it. &Instead of making better choices, we'll \textcolor{blue}{be} overwhelmed by choice, sometimes even afraid. &Instead of meeting better decisions, we \textcolor{blue}{get} overwhelmed by choice, sometimes it makes us fearful. & Instead of the meeting of better choices, we\textcolor{blue}{'re} even afraid of choice. \\
\midrule
\midrule
fresh &Wir reden hier über gute, frische Lebensmittel, die in unglaublichem Ausmaß verschwendet werden. &We're talking about good, \textcolor{red}{fresh} food that is being wasted on a colossal scale. &We're talking about good, \textcolor{red}{fresh} food that's being used in incredible scale. &We're talking about good, \textcolor{blue}{new} food that's used in incredible scale. &We're talking about good foods that's going to be used in the incredible order of scale. \\
\midrule
fresh &Wir brauchen einen neuen Standard für ordentliches frisches Essen für eure Kinder. Ja? &There needs to be a new standard of \textcolor{red}{fresh}, proper food for your children. Yeah? &We need a new standard for decent \textcolor{red}{fresh} food for your kids. Yes? &We need a new standard for proper \textcolor{blue}{new} food for your kids. Right? &We need a new set of \textcolor{blue}{clean} food for your kids. Yes? \\
\midrule
fresh &Ich glaube dass hier zwei frische Ideen drin sind -- zwei. &Well, I think there are two \textcolor{red}{fresh} things here -- two fresh things. &I think there's two \textcolor{red}{fresh}est ideas in here -- two fresh water. &I think there are two \textcolor{blue}{new} ideas in it -- two. &I believe that there's two \textcolor{blue}{new} ideas -- two. \\
\midrule
\midrule
energy &Sie werden unübertroffene Vitalität und Energie gewinnen. &You'll have unsurpassed vitality and \textcolor{red}{energy}. &They're unsurprising vitality and \textcolor{red}{energy}. &They're won't win overblown vitality and \textcolor{blue}{power}. &It's become overconducted vitality and \textcolor{blue}{power}. \\
\midrule
energy &Also habe ich gedacht, wie wir die Energiekrise in diesem Land bewältigen können? &And so I thought, how could we address the \textcolor{red}{energy} crisis in this country? &So I thought, how do we deal with the \textcolor{red}{energy} crisis in this country? &So I thought, how can we deal with the \textcolor{blue}{power} crisis in this country? &So I thought, how do we deal with the crisis in this country? \\
\midrule
energy &Energiepflanzen liefern ein halbes Watt pro Quadratmeter in europäischem Klima. &\textcolor{red}{Energy} crops deliver half a watt per square meter in European climates. &\textcolor{red}{Energy} crops deliver half a watt per square meter in European climates. &\textcolor{blue}{Power} plants provide half watts per square meter in the European climate. &\textcolor{blue}{power} plants deliver half a watt per square meter in European climates. \\
\bottomrule
\end{tabular}
}
\end{center}
\vskip -0.5em
\caption{\label{appendix:tab:case1} 
Translation cases before and after unlearning from IWSLT dataset. After removing all training instances containing specific words (in \textcolor{red}{red}), the unlearned models tend to generate alternatives (in \textcolor{blue}{blue}) with similar meanings to maintain consistency in terms of the whole sentence. For example, after removing all instances containing ``\textcolor{red}{energy}'' from original training set, the unlearned model generates ``\textcolor{blue}{power}'' in corresponding positions. 
}
\vskip -0.5em
\end{table*}

\paragraph{Parameter Setting and Training.}
Apart from the brief description in \Cref{sec:exp:setup}, we give more experimental details here. 
The DistilBERT we used for LEDGAR contains 6 transformer encoder layers each with 768 dimensions and 3072-dimensional feed-forward networks, resulting in 67M parameters. The transformer models used for IWSLT and PersonaChat are of the same size, i.e., containing 6 encoder and decoder layers each with 512 dimensions and 1024-dimensional feed-forward networks, with a total parameter amount of 91M.
For the LSTM and BART-Base models we use in \Cref{sec:exp:further}, the model sizes are 40M and 251M, respectively. The LSTM model contains 2 layers of encoder and decoder respectively, with 512 hidden size. The BART-Base model has 6 layers of 768-dimensional encoder and decoder, where we follow \citet{lewis-etal-2020-bart} to add new sets of encoder parameters before the pretrained BART encoder. This results in total 10 encoder layers (i.e., we add 4 layers).

We use one NVIDIA RTX 3090 GPU to train our model.
When training the original model, the batch size is selected from $\{$16, 32, 64$\}$, and the final choices are 32 for LEDGAR and IWSLT, and 16 for PersonaChat, with an update frequency of 8. Learning rate is selected in $\{$1e-3, 5e-4, 2e-4, 5e-5, 2e-5$\}$, and we use 5e-5 for LEDGAR, 5e-4 for IWSLT, and 2e-4 for PersonaChat, respectively. Dropout strategy~\cite{Srivastava:2014:DSW:2627435.2670313} with dropout rate selected in $\{$0.1, 0.2, 0.3$\}$ (the final choice is 0.1 for LEDGAR, and 0.3 for IWSLT and PersonaChat) and $L_2$ regularization with 0.0001 effect value are used to alleviate overfitting. During inference in generation tasks, the beam size is set to 5. All the above hyper-parameters are selected based on the performance of validation set.


\paragraph{Unlearning Setting.}
The removal numbers are set to 100 instances for LEDGAR and IWSLT, and 10 conversations (about 100 instances) for PersonaChat unless otherwise noted. We set the stopping hyper-parameter $\sigma$ to $0.1$. 

\section{More Translation Unlearning Cases}
\label{appendix:more_case}
\Cref{appendix:tab:case1} shows more cases when deleting all instances containing specific words, including ``become'' (verb), ``fresh'' (adjective), and ``energy'' (noun). We can find that unlearned models (i.e., \textsc{Retrain} and \textsc{Kga}) tend to generate alternatives with similar meanings regardless of the part of speech.

\end{document}